# System-Level Error Propagation and Tail-Risk Amplification in Reference-Based Robotic Navigation

Ning Hu[1*], Maochen Li[2], Senhao Cao[3]

[1] Department of Mechanical, Aerospace and Biomedical Engineering, University of Tennessee, Knoxville, TN 37916, USA

[2] ZOEZEN ROBOT CO.LTD, Beijing, China

[3] Northeastern University, Boston, MA 02115, USA

*Correspondence: nhu2@vols.utk.edu

*Abstract*— Image guided robotic navigation systems frequently rely on reference-based geometric perception pipelines, where accurate spatial mapping is established through multi-stage estimation processes. In biplanar X-ray–guided navigation, such pipelines are widely adopted due to their real-time capability and geometric interpretability. However, navigation reliability is often constrained by an overlooked system-level failure mechanism in which installation-induced structural bias introduced at the perception stage can be progressively amplified along the perception–reconstruction–execution chain and ultimately dominate execution-level error and tail-risk behavior. This paper investigates this mechanism from a system-level perspective and presents a unified error propagation modeling framework that explicitly characterizes how installation-induced structural bias propagate and couple with pixel-level observation noise through biplanar imaging, projection matrix estimation, triangulation, and coordinate mapping. By combining first order analytic uncertainty propagation with large-scale Monte Carlo simulations, we analyze dominant sensitivity channels and quantify worst-case error behavior beyond mean accuracy metrics. The results demonstrate that rotational installation error acts as a primary driver of system level error amplification, whereas translational misalignment of comparable magnitude plays a secondary role under typical biplanar geometries. Moreover, installation-induced structural bias fundamentally alter the system's sensitivity to perception noise, leading to pronounced tail-risk amplification that cannot be captured by additive error models. Real biplanar X-ray bench-top experiments further confirm that the predicted error amplification trends persist under realistic imaging and feature extraction uncertainty. Beyond the specific context of biplanar X-ray navigation, the findings reveal a broader structural limitation of reference-based, multi-stage geometric perception pipelines in robotics. By explicitly modeling system-level error propagation and tail-risk behavior, this work formulates a systematic framework for reliability assessment, sensitivity analysis, and risk-aware design in safety-critical robotic navigation systems.

*Keywords-System-level error propagation, Tail-risk amplification, Reference-based navigation, Geometric perception pipelines, Multi-stage geometric estimation, Uncertainty propagation, Sensitivity analysis, Biplanar X-ray navigation, Safety-critical robotic systems*

## I. Introduction

Robotic systems are increasingly deployed in safety-critical scenarios that require precise geometric reasoning under stringent reliability constraints, including image-guided intervention, surgical navigation, and reference-based robotic manipulation [1-3]. In such systems, perception outputs are mapped to physical execution through a chain of geometric transformations, and even millimeter-level deviations may lead to irreversible consequences. As a result, many robotic platforms adopt reference-based navigation pipelines to establish spatial consistency between sensing and actuation [4-5].

A representative class of these systems relies on multi-stage geometric perception pipelines, in which spatial relationships are inferred through sequential estimation steps, including image formation, projection matrix estimation, geometric reconstruction, and coordinate mapping [6-9]. Such pipelines are widely used due to their interpretability and real-time capability. In practice, reliability is often assumed once local calibration accuracy and sensor noise are sufficiently controlled.

However, this assumption reflects a component-level uncertainty paradigm. In many real robotic systems, execution-level deviations remain large, anisotropic, and directionally biased even when perception metrics appear well controlled[10-13]. This discrepancy suggests that system-level behavior cannot be fully explained by independent noise sources or local calibration error alone. Instead, error may propagate and interact across the entire perception–reconstruction–execution chain in a structurally coupled manner.

A particularly underexplored source of uncertainty arises from **installation-induced structural bias**—small but systematic misalignments in reference configurations that are ubiquitous in practical robotic deployments. Unlike random measurement noise, **this structural bias** perturbs the assumed geometry in a systematic manner and cannot be eliminated through calibration refinement. Whether such structural bias can be systematically amplified through nonlinear geometric estimation, and under what conditions it dominates worst-case execution-level reliability, remains insufficiently understood.

From a system-level perspective, this paper investigates the following question:

Under what conditions can small structural bias in a reference-based geometric configuration be amplified into disproportionately large execution-level errors, even when local calibration accuracy is high?

We refer to this phenomenon as system-level structural error amplification, where upstream structural bias is amplified into execution-level tail risk through a multi-stage geometric estimation chain.

We argue that this phenomenon reflects an inherent structural limitation of reference-based, multi-stage geometric perception pipelines. Specifically, installation-induced structural bias alters the three-dimensional control point configuration used in projection matrix estimation and is subsequently amplified through nonlinear reconstruction and coordinate mapping processes. The resulting execution error exhibits pronounced tail-risk behavior that cannot be captured by mean accuracy metrics alone.

To analyze this mechanism in a concrete yet representative setting, we study biplanar X-ray–based 3D localization as an instance of reference-driven robotic perception pipelines[6-8]. While such systems are often assumed to be robust under controlled calibration, we demonstrate that even small installation misalignments—particularly rotational perturbations—can fundamentally reshape system sensitivity and dominate execution-level reliability through coupled amplification across estimation stages.

Motivated by this observation, we develop a unified system-level error propagation framework spanning projection matrix estimation, 3D reconstruction, and coordinate mapping. Grounded in uncertainty propagation theory, the framework explicitly models how structural bias interacts with pixel-level noise and alters sensitivity channels in the estimation chain. Our analysis reveals a pronounced sensitivity asymmetry: rotational installation error emerges as the dominant amplification channel governing both localization accuracy and tail-risk behavior, whereas translational misalignment of comparable magnitude plays a secondary role under typical geometric configurations.

The proposed framework is validated through first-order analytic uncertainty propagation, large-scale Monte Carlo simulation, and real-system phantom experiments. Rather than serving as performance benchmarks, these experiments are designed as mechanism validation studies, demonstrating consistency between analytic prediction, statistical characterization, and observed execution-level behavior.

Importantly, this work does not introduce a new navigation algorithm or calibration procedure. Instead, it exposes a structural error amplification mechanism inherent in reference-based robotic perception pipelines. By explicitly modeling system-level error propagation and tail-risk emergence, this study establishes a principled foundation for reliability assessment, sensitivity analysis, and risk-aware design in safety-critical robotic systems.

The main contributions of this work are threefold:

1. We identify and formalize a structural error amplification mechanism in reference-based, multi-stage geometric perception pipelines.
2. We quantitatively characterize the coupled interaction between installation-induced structural bias and perception noise, demonstrating that execution-level error behavior is not a linear superposition of independent uncertainty sources.
3. We establish a unified validation framework linking analytic modeling, statistical characterization, and real-system experimentation, clarifying the predictive scope and limitations of first-order uncertainty propagation in nonlinear robotic estimation chains.

## II. RELATED WORK

This section reviews prior research through the lens of uncertainty modeling and robotic reliability, highlighting a fundamental gap between component-level accuracy analysis and system-level structural amplification in reference-based robotic pipelines.

### A. Local Error Modeling in Geometric Reconstruction

Classical studies in multi-view geometry and image-guided navigation have established rigorous foundations for projection matrix estimation, triangulation, and 3D reconstruction [6–9]. In medical navigation contexts, extensive research has analyzed target registration error (TRE), reprojection error, and the influence of imaging geometry on reconstruction fidelity [10][20–22]. These works have significantly advanced the understanding of how measurement noise propagates through geometric estimation processes.

However, most existing analyses adopt a local uncertainty modeling perspective. Deviations in control point coordinates or pixel observations are typically treated as independent random perturbations, and error behavior is characterized using linearized covariance propagation or mean/variance metrics. While such approaches provide valuable insight into noise sensitivity, they implicitly assume that the underlying geometric configuration remains unbiased. The possibility that small but systematic structural bias in reference configurations may alter the conditioning of the estimation chain and fundamentally reshape downstream execution-level reliability remains largely unexplored.

### B. Calibration Chains and Component-Wise Accuracy Optimization

Robotic navigation systems rely on a sequence of extrinsic transformations, including hand–eye calibration, TCP definition, and coordinate frame alignment [9][17–19]. A substantial body of work has focused on improving calibration accuracy, repeatability, and robustness through refined optimization strategies and marker-based or markerless methods.

These studies have been highly effective in reducing individual transformation error. However, they predominantly analyze each calibration component in isolation. In practice, robotic execution accuracy emerges from the composition of multiple geometric transformations and reconstruction stages. Structural deviations introduced upstream—such as installation-induced misalignment of reference structures—may interact with calibration errors in a non-additive manner. Existing calibration literature does not explicitly model how such structural bias propagates across the perception–reconstruction–execution chain or how it influences worst-case execution-level reliability. This implicitly assumes that error sources combine linearly, an assumption that breaks down under structurally biased geometric inputs.

### C. Execution-Level Robustness and Reliability Assessment

In safety-critical robotic systems, reliability assessment has increasingly emphasized task-level accuracy and execution robustness [1–5]. In image-guided intervention and surgical navigation, research has moved beyond landmark fitting accuracy toward target point accuracy as a more meaningful metric [10][20]. Nonetheless, reliability evaluation remains largely centered on mean error, variance, or deterministic worst-case bounds under simplified assumptions.

What remains insufficiently characterized is the emergence of tail-risk–dominated behavior resulting from structurally biased geometric inputs. When installation-induced structural bias alters the effective reference configuration, nonlinear geometric estimation may amplify upstream bias into execution-level anisotropy and heavy-tailed error distributions. Such coupled amplification cannot be captured by component-wise accuracy improvement or linear superposition of independent uncertainty sources.

### D. Summary and Gap

In summary, prior research has established rigorous foundations in geometric reconstruction, calibration, and uncertainty propagation. However, these studies largely operate within a component-level uncertainty paradigm, where individual error sources are analyzed and minimized in isolation. Such a perspective implicitly assumes linear superposition and unbiased geometric configurations.

A unified system-level framework that models installation-induced structural bias as systematic bias in geometric inputs—and that characterizes their nonlinear amplification and tail-risk emergence across the entire perception–reconstruction–execution chain—remains absent. Addressing this gap requires shifting from component-wise accuracy optimization to structural reliability modeling in reference-based robotic systems.

This paper responds to this need by revealing and formalizing a structural error amplification mechanism inherent in multi-stage geometric perception pipelines. Rather than proposing incremental algorithmic refinements, we provide a principled system-level explanation for how small upstream structural bias can dominate downstream execution-level reliability.

## III. METHODOLOGY

Fig. 1 illustrates the system-level error propagation pipeline analyzed in this work. The navigation system follows a representative reference-based geometric perception workflow: a known three-dimensional reference structure is observed by a biplanar C-arm X-ray imaging system, reconstructed through multi-stage geometric estimation, and mapped to the robot tool center point (TCP) for execution.

While such pipelines are commonly regarded as reliable once local calibration accuracy and imaging noise are controlled, this assumption implicitly adopts a component-level uncertainty paradigm. In contrast, we analyze how installation-induced structural bias modifies the geometric configuration itself and reshapes the sensitivity structure of the entire perception–reconstruction–execution chain.

In this section, we formalize this mechanism and establish a unified framework for analyzing **system-level structural error amplification**. Detailed derivations are provided in the Supplementary Material.

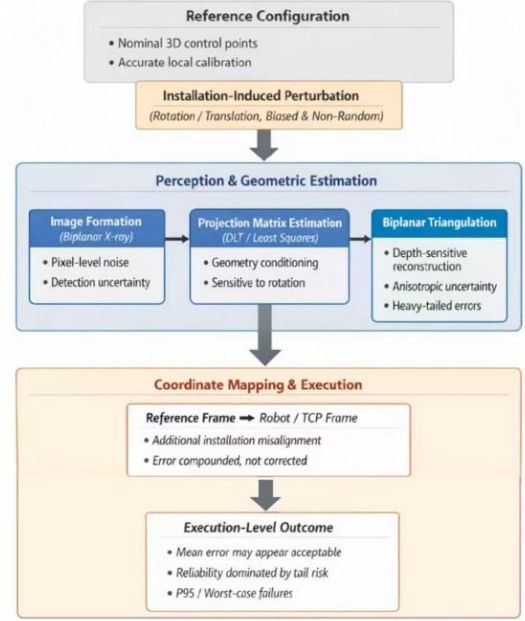

Fig 1. Overview of system-level error propagation in reference-based geometric perception pipelines. Installation-induced structural bias introduce biased geometric inputs at the perception stage, which are nonlinearly amplified through projection matrix estimation, biplanar triangulation, and coordinate mapping. This amplification fundamentally alters system sensitivity to perception noise and leads to heavy-tailed execution-level error distributions, where tail risk rather than mean accuracy dominates navigation reliability.

### A. Imaging and Localization Baseline

In biplanar X-ray–guided robotic navigation, a rigid three-dimensional reference structure attached to the robot is observed from two distinct imaging views. Two-dimensional projections of reference features are associated with their nominal three-dimensional coordinates to estimate view-dependent projection matrices. Target points are reconstructed via biplanar triangulation and subsequently mapped through a sequence of coordinate transformations to the robot TCP frame for execution.

This process can be abstracted as a composition of nonlinear operators:

projection estimation → triangulation → coordinate mapping → execution.

Under nominal conditions, the geometric configuration of the reference structure is assumed to be accurate. Let this nominal configuration be denoted as $G_0$. In the absence of structural bias, uncertainty arises primarily from pixel-level observation noise, and system behavior is governed by the local sensitivity of these nonlinear operators evaluated at $G_0$.

This nominal formulation serves as the baseline against which structural bias–induced deviations are analyzed.

In biplanar X-ray–guided robotic navigation, a known three-dimensional reference structure rigidly attached to the robot is observed from two distinct imaging views. Two-dimensional projections of reference features are associated with their nominal three-dimensional coordinates to estimate view-dependent projection matrices, and target points are reconstructed via biplanar triangulation. The reconstructed target position is then mapped through a sequence of coordinate transformations to the robot tool center point (TCP) frame for execution. This reference-based, multi-stage geometric pipeline is widely adopted due to its real-time capability and geometric interpretability. In this work, the formulation is treated as a baseline, and emphasis is placed on how deviations in the assumed reference geometry propagate through this pipeline and affect execution-level reliability.

### B. Structural Bias and Operator Conditioning

We model installation-induced structural bias as a small but systematic deviation in the assumed geometric configuration of the reference structure prior to projection matrix estimation. Let the effective configuration be denoted abstractly as:

$$G = G_0 + \Delta G,$$

Where $\Delta G$ represents installation-induced structural bias.

In the Supplementary Material, this abstract representation corresponds to perturbed transformation matrices introduced by installation error (e.g., $T_{err1}, T_{err2}$), which modify the nominal transformation $A_{true}$ and yield an effective transformation $A_{actual}$. In this sense, structural bias does not act as additive measurement noise; rather, it alters the geometric model itself.

Crucially, the nonlinear operators involved in projection estimation and triangulation are geometry-dependent. Their Jacobians are evaluated with respect to the underlying geometric configuration. Therefore, when the system operates under $G = G_0 + \Delta G$, the effective Jacobians differ from those evaluated at $G_0$. Structural bias thus modifies the conditioning of the estimation chain.

This change in conditioning constitutes the core mechanism of system-level structural error amplification. Instead of simply adding error, structural bias reshapes the sensitivity landscape of the operators, thereby amplifying downstream deviations.

In particular, rotational misalignment globally distorts the spatial distribution of control points, significantly altering the conditioning of projection estimation and triangulation. This leads to amplified sensitivity to pixel-level noise and induces anisotropic error growth. In contrast, translational deviations of comparable magnitude do not modify geometric conditioning to the same extent under typical biplanar configurations, resulting in secondary effects.

The overall execution error can therefore be interpreted as the outcome of a composed nonlinear mapping in which upstream structural bias modifies operator conditioning, and subsequent stages propagate and amplify this modification.

### C. Uncertainty Propagation and Tail-Risk Emergence

To quantify system-level structural error amplification, we analyze uncertainty using two complementary approaches: first-order analytic propagation and Monte Carlo simulation.

First-order analytic propagation captures dominant sensitivity directions and relative scaling relationships under local linearization. In particular, because the Jacobians depend on the effective geometry $G$, structural bias alters these Jacobians and therefore modifies the mapping from pixel-level noise to execution-level error. This provides interpretable insight into how structural bias reshapes the sensitivity structure of the estimation chain.

Monte Carlo simulation complements this analysis by capturing higher-order nonlinear effects. Under biased geometric configurations, the curvature of the nonlinear mappings increases, leading to skewed and heavy-tailed error distributions in execution space.

Importantly, the combined effect of structural bias and pixel-level noise is not additive. Structural bias modifies operator conditioning, introducing multiplicative coupling between geometric configuration and measurement noise. As a result, otherwise tolerable perception noise can be amplified into execution-level tail risk.

Accordingly, navigation reliability is evaluated not only through mean localization accuracy but also through tail-risk metrics, such as high-percentile execution error. These metrics more faithfully reflect worst-case behavior in safety-critical robotic systems and reveal the dominance of structural bias in shaping execution-level reliability.

## IV. EXPERIMENTS AND ANALYSIS

### A. Experimental Objectives and Validation Logic

The experiments are designed to validate the system-level error propagation mechanism induced by installation-induced structural bias. Specifically, we aim to answer three questions: (1) where structural errors enter the perception–reconstruction–execution pipeline and how they are amplified; (2) whether installation errors couple with perception noise in a non-additive manner; and (3) whether the proposed model captures dominant sensitivity channels and tail-risk behavior. To this end, two simulation studies are conducted to isolate the primary and coupled effects of installation error, followed by a minimal physical phantom experiment to verify that the observed mechanisms persist under real imaging conditions. These experiments are designed not merely to observe error growth, but to empirically validate the structural conditioning change and non-additive amplification mechanism formalized in Section III.

### B. Simulation Study I: Primary Effect of Installation Error

We first examine whether installation-induced structural bias constitute the dominant driver of system-level error amplification. Pixel-level noise is fixed at a negligible level, while rotational and translational installation errors are independently varied to isolate their respective effects on localization and execution accuracy.

As shown in Fig. 2, both the 3D localization error and the TCP execution error increase monotonically with rotational installation misalignment. Increasing the rotation error from 0° to 2° leads to a substantial growth in both mean error and tail-risk metrics, indicating pronounced amplification of execution-level deviation. In this experiment, the TCP transformation is assumed to be an ideal rigid-body mapping without additional execution uncertainty. Consequently, execution error closely matches localization error magnitude, demonstrating that structural bias introduced at the perception and reconstruction stages alone are sufficient to determine the upper bound and tail-risk behavior of navigation accuracy.

In contrast, translational installation error produces only marginal variations in both localization and execution error over the tested range, as illustrated in Fig. 3. This sharp asymmetry confirms that, under the tested biplanar geometry and target scale, rotational misalignment constitutes the dominant structural error source, whereas translational deviations of comparable magnitude act primarily as secondary perturbations. This observed asymmetry is consistent with the conditioning analysis presented in Section III, where rotational structural bias significantly modifies the effective Jacobian structure of the estimation operators, whereas translational deviations induce comparatively minor conditioning shifts under typical geometric configurations. This distinction highlights the fundamentally different roles of rotational and translational installation errors in shaping system-level sensitivity and reliability. Quantitative results for simulation I configurations, including full error statistics and tail-risk metrics, are summarized in the Supplementary Material.

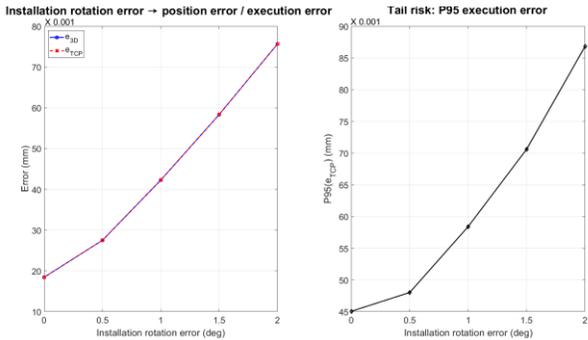

Fig.2. Effect of installation rotation error. Left: localization and execution errors increase approximately linearly with rotation misalignment. Right: tail risk (P95) of execution error grows significantly with installation error.

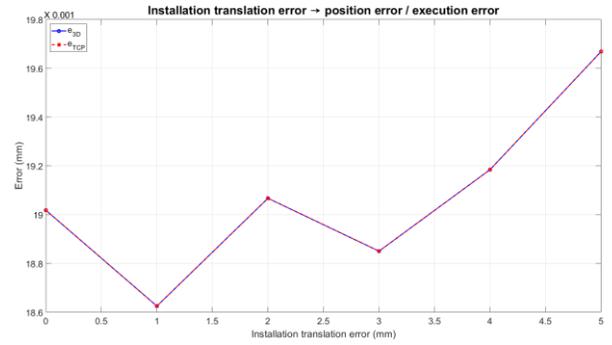

Fig.3. Effect of installation translation error. Translation misalignment introduces relatively minor perturbations compared to rotational errors under the tested geometry.

### C. Simulation Study II: Coupled Amplification of Installation Error and Pixel Noise

In practical robotic navigation systems, installation-induced structural bias and pixel-level perception noise coexist. To examine their interaction, we conduct a coupled simulation study in which installation error levels and pixel noise magnitude are jointly varied, and execution-level tail-risk metrics are evaluated.

As shown in Fig. 4, pixel noise and installation-induced structural bias exhibit a pronounced non-additive coupling effect. In the absence of installation error, increasing pixel noise leads to a relatively bounded growth in execution error. Once installation-induced structural bias are introduced, however, the same level of pixel noise produces a substantially larger increase in tail risk. This sharp change in sensitivity indicates that structural bias modifies the effective estimation operators themselves, introducing multiplicative coupling between geometric configuration and pixel-level noise, rather than simple additive error accumulation.

These results demonstrate that system-level error cannot be explained as a simple superposition of independent noise sources. Instead, installation-induced structural bias alter the geometric conditioning of the estimation pipeline, transforming otherwise tolerable perception noise into a navigation-level risk factor through coupled amplification.

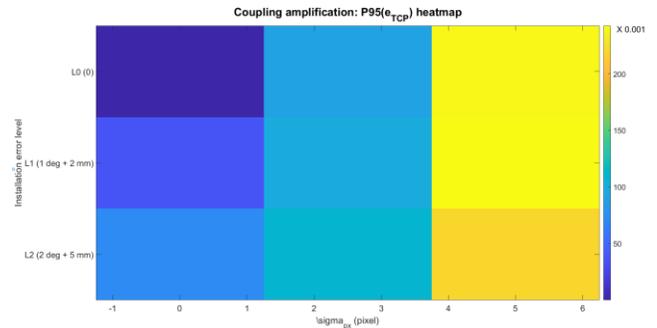

Fig.4. Coupled amplification of pixel noise and installation error. Tail risk (P95) of execution error increases sharply with pixel noise under different installation error levels.

### D. Physical Phantom Experiment

To validate the simulation findings at the execution level, a minimal biplanar X-ray phantom experiment is conducted under controlled installation conditions. The reference structure is installed with increasing levels of misalignment, and biplanar images are repeatedly acquired to estimate localization error and tail-risk metrics.

Experimental results exhibit a clear monotonic increase in error magnitude and tail risk with increasing installation perturbation, consistent with simulation predictions in both trend and relative magnitude. These results confirm that the identified system-level error amplification mechanism persists under realistic imaging and detection uncertainty. The monotonic growth in both mean error and tail-risk metrics further supports the theoretical prediction that structural bias reshapes operator conditioning and amplifies nonlinear estimation effects in execution space. Quantitative results for simulation II configurations, including full error statistics and tail-risk metrics, are summarized in the Supplementary Material.

To further assess the predictive scope and limitations of first-order analytic uncertainty propagation, we compare analytic predictions with Monte Carlo statistics under a representative operating condition. Detailed results and discussion are provided in the Supplementary Material.

*E. Summary*

Across simulation and real-system validation, the results establish a coherent evidence chain demonstrating that installation-induced structural bias act as a dominant driver of system-level error amplification. The observed coupled interaction between structural bias and perception noise further highlights the limitation of local accuracy metrics and motivates a shift from local accuracy optimization toward end-to-end, system-level reliability analysis in reference-based geometric navigation systems.

## V. REAL-SYSTEM VALIDATION

To examine whether the identified system-level error amplification mechanism persists under real imaging conditions, we conduct a biplanar X-ray bench-top experiment. This experiment is not intended as a clinical evaluation, but as a mechanism validation to assess whether the error propagation behaviors revealed by simulation remain observable in a real navigation pipeline subject to imaging noise and feature extraction uncertainty.

As shown in Fig. 5, a C-arm X-ray system acquires anteroposterior (AP) and lateral (LAT) views of a localization reference structure mounted to a rigid phantom. Due to practical engineering constraints, rotational and translational installation errors cannot be independently controlled. Instead, a level-based installation strategy is adopted, defining three representative conditions: ideal installation (L0), moderate misalignment (L1), and larger misalignment (L2). Each condition is repeated with full disassembly and reinstallation to ensure that observed variability reflects installation-induced effects.

Across installation levels, both localization error and tail-risk metrics exhibit a clear monotonic increase, consistent with simulation predictions in trend and relative magnitude. Importantly, beyond quantitative metrics, the execution-level outcomes shown in Fig. 6 reveal visually apparent and directionally consistent degradation from L0 to L2, despite identical target definitions and nominal perception accuracy. This behavior indicates a structurally biased mapping rather than noise-induced fluctuation.

These real-system results provide direct qualitative and quantitative evidence that small installation-induced structural bias can be systematically amplified through the perception–reconstruction–execution chain. The observed trends closely mirror those identified in simulation, confirming that system-level error amplification and tail-risk-dominated behavior persist under realistic imaging conditions. Quantitative localization and reprojection error statistics for all installation conditions are summarized in the Supplementary Material.

Together, these findings provide empirical validation of the system-level structural error amplification mechanism defined in Section III, demonstrating that upstream structural bias systematically dominates execution-level reliability in real robotic navigation systems.

A representative execution-level demonstration is provided in the supplementary video, illustrating the monotonic and directionally consistent degradation under increasing structural bias.

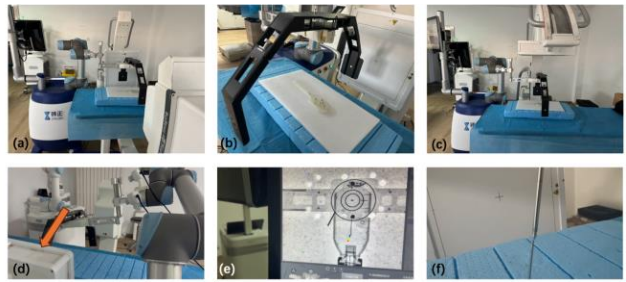

Fig.5 Real-system biplanar X-ray navigation setup and experimental configuration for system-level error propagation analysis. (a–c) Biplanar X-ray bench-top experimental setup under different installation states, illustrating the physical configuration used for real-system validation. (d) Geometric workflow of the biplanar X-ray–guided navigation pipeline, including reference-based projection matrix estimation, biplanar triangulation, and coordinate mapping to the robot execution frame. (e) Representative anteroposterior (AP) and lateral (LAT) X-ray views used for geometric localization. (f) Illustration of the execution task and positioning error definition following target insertion. Together, these components define the perception–reconstruction–execution chain through which installation-induced structural bias enter and propagate in the real system.

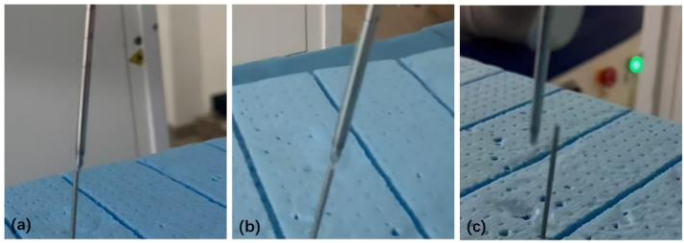

Fig. 6. Execution-level manifestation of system-level error amplification under increasing installation-induced structural bias. (a–c) Final execution outcomes under three installation conditions: ideal installation (L0), moderate misalignment (L1), and larger misalignment (L2), corresponding to increasing levels of installation-induced structural perturbation. Despite identical target definitions and nominal perception accuracy, the execution result exhibits a visually apparent and monotonic degradation from L0 to L2. The observed deviation is directionally consistent across trials, indicating a structurally biased mapping rather than noise-induced fluctuation. These execution-level outcomes

provide qualitative but direct evidence that small structural bias introduced at the perception stage can be systematically amplified through the perception–reconstruction–execution chain, leading to tail-risk-dominated failures.

## VI. Conclusion

This paper examined a system-level error propagation mechanism in reference-based geometric perception pipelines, with particular emphasis on installation-induced structural bias. We formalized this phenomenon as **system-level structural error amplification**, in which upstream structural bias modifies the conditioning of multi-stage geometric estimation operators and is subsequently amplified into execution-level tail risk. Rather than treating uncertainty as independent noise confined to individual modules, our analysis demonstrated that structural bias reshapes the sensitivity structure of the entire perception–reconstruction–execution chain.

Through complementary simulation studies, we showed that rotational installation error constitutes the dominant driver of structural error amplification under typical biplanar geometries, whereas translational misalignment of comparable magnitude plays a secondary role. First-order analytic uncertainty propagation captured dominant sensitivity directions and relative scaling behavior, while Monte Carlo analysis revealed nonlinear amplification and heavy-tailed execution error distributions. Together, these results clarified both the operator-level conditioning changes induced by structural bias and the limitations of purely linearized error models.

Real-system biplanar X-ray experiments further confirmed that the identified amplification mechanism persists under realistic imaging and feature extraction uncertainty. The monotonic growth in mean localization error and P95 tail-risk metrics across installation levels, together with visually apparent and directionally consistent execution degradation, demonstrated that structural amplification is not a simulation artifact but manifests as tangible execution-level reliability failure in physical robotic systems.

Beyond the specific context of biplanar X-ray–guided navigation, these findings reveal a broader structural limitation of reference-based, multi-stage geometric perception pipelines in robotics. Small structural bias introduced early in the estimation chain can systematically dominate downstream execution-level reliability through conditioning shifts and nonlinear amplification. This observation motivates a shift from component-level accuracy optimization toward system-level structural reliability modeling in safety-critical robotic systems.

Future work will extend this framework to incorporate execution-side uncertainties, such as hand–eye calibration error and manipulator compliance, and to explore risk-aware calibration, planning, and control strategies informed by system-level structural error amplification.

# Supplementary

## I. METHODOLOGY

### A. Imaging and Localization Overview

In medical image-guided navigation, C-arm X-ray imaging can be abstracted as a pinhole camera system. The pinhole camera model is one of the most widely used geometric models in computer vision, describing the mapping from three-dimensional space to the two-dimensional image plane through linear projection. Its core lies in estimating the intrinsic and extrinsic parameters of the imaging system.

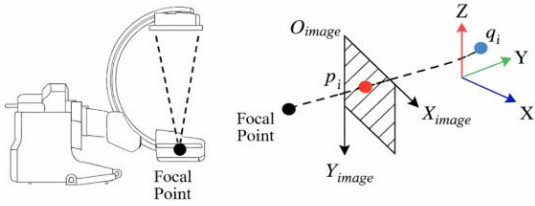

Fig. S1 Overview of Camera Imaging Model and Localization

As illustrated in Fig. S1, the X-ray pinhole camera model maps a 3D feature point expressed in the object coordinate system to the X-ray image coordinate system via a projective transformation. The mapping equation is given by:

$$\mathbf{p} = \mathbf{A}\mathbf{q}$$
$$\mathbf{p} = s[u \ v \ 1]^\top,$$
$$\mathbf{A} = \begin{bmatrix} A_1 & A_2 & A_3 & A_4 \\ A_5 & A_6 & A_7 & A_8 \\ A_9 & A_{10} & A_{11} & A_{12} \end{bmatrix}, \quad (1)$$
$$\mathbf{q} = [x \ y \ z \ 1]^\top$$

Where $\mathbf{p}$ denotes the homogeneous 2D coordinates of a feature point in the X-ray image coordinate system, $\mathbf{q}$ denotes the homogeneous 3D coordinates of the corresponding feature point in the object coordinate system, $s$ is a scale factor, $[u \ v \ 1]^\top$ represents the image coordinates, $\mathbf{A}$ is the transformation (projection) matrix, and $[x \ y \ z \ 1]^\top$ denotes the 3D coordinates in the target coordinate system. The transformation matrix $\mathbf{A}$ contains 12 unknown parameters. Substituting the coordinates into the mapping equation yields:

Here $\mathbf{x}_i$ denotes the homogeneous 2D image coordinates of the feature point in the X-ray image, $\tilde{\mathbf{X}}_i$ represents the homogeneous 3D coordinates of the corresponding feature point in the object coordinate system, $\lambda$ is a scale factor, and $\mathbf{P}$ is the projection matrix. The projection matrix $\mathbf{P}$ contains 12 unknown parameters and encapsulates both intrinsic and extrinsic imaging geometry.

$$\begin{aligned}(A_1 - A_9 u)x + (A_2 - A_{10}u)y + (A_3 - A_{11}u)z + (A_4 - A_{12}u) = 0 \\ (A_5 - A_9 v)x + (A_6 - A_{10}v)y + (A_7 - A_{11}v)z + (A_8 - A_{12}v) = 0\end{aligned} \quad (2)$$

That is, $G_u(\hat{\mathbf{A}}, \mathbf{p}, \mathbf{q}) = 0$ and $G_v(\hat{\mathbf{A}}, \mathbf{p}, \mathbf{q}) = 0$. Using $n$ point correspondences, a system of $2n$ nonlinear equations can be constructed, i.e.,

$$G(\tilde{\mathbf{A}}, \boldsymbol{\alpha}) = 0 \quad (3)$$

Where $\tilde{\mathbf{A}} = [A_1, A_2, \ldots, A_{12}]^\top$, $\boldsymbol{\alpha} = [\boldsymbol{\alpha}_1, \boldsymbol{\alpha}_2, \ldots, \boldsymbol{\alpha}_n]^\top$, $\boldsymbol{\alpha} = [x, y, z, u, v]$, (3) can be solved using a least-squares formulation.

To achieve three-dimensional localization of a target point, the system requires X-ray images acquired from at least two distinct viewing angles. In practice, a biplanar configuration is typically adopted, consisting of anteroposterior (AP) and lateral (LAT) views. Under ideal, noise-free conditions, two projection matrices can be estimated by solving Eq. (3) for each view. Given a corresponding point observed in both images, its three-dimensional position can then be reconstructed through the intersection of the associated projection rays.

$$\mathbf{p}_1 = \mathbf{A} \cdot \mathbf{q}_{\text{aim}}, \quad \mathbf{p}_2 = \mathbf{B} \cdot \mathbf{q}_{\text{aim}} \quad (4)$$

$$s\begin{bmatrix} u_1 \\ v_1 \\ 1 \end{bmatrix} = \begin{bmatrix} A_{1,:} \\ A_{5,:} \\ A_{9,:} \end{bmatrix}\begin{bmatrix} x \\ y \\ z \\ 1 \end{bmatrix} \quad s\begin{bmatrix} u_2 \\ v_2 \\ 1 \end{bmatrix} = \begin{bmatrix} B_{1,:} \\ B_{5,:} \\ B_{9,:} \end{bmatrix}\begin{bmatrix} x \\ y \\ z \\ 1 \end{bmatrix} \quad (5)$$

$$\left(u_1 A_{9,:} - A_{1,:}\right)\begin{bmatrix} x \\ y \\ z \\ 1 \end{bmatrix} = 0 \quad \left(v_1 A_{9,:} - A_{5,:}\right)\begin{bmatrix} x \\ y \\ z \\ 1 \end{bmatrix} = 0$$

$$\left(u_2 B_{9,:} - B_{1,:}\right)\begin{bmatrix} x \\ y \\ z \\ 1 \end{bmatrix} = 0 \quad \left(v_2 B_{9,:} - B_{5,:}\right)\begin{bmatrix} x \\ y \\ z \\ 1 \end{bmatrix} = 0 \quad (6)$$

$$N \cdot \mathbf{q}_{\text{aim}} = 0 \quad (7)$$

The matrix $\mathbf{N}$ is decomposed using singular value decomposition (SVD), $\mathbf{q}_{\text{aim}} = \text{SVD}(N)$, $N = U\Sigma V^\top$, The solution vector $\mathbf{q}_{\text{aim}}$ is given by the right singular vector associated with the smallest singular value, i.e., the last column of $\mathbf{V}$. After normalization, $\mathbf{q}_{aim}$ represents the reconstructed three-dimensional coordinates of the target point.

*B. Structural Error Propagation Mechanism*

Error Analysis

In practical image-guided surgical applications, various sources of uncertainty are inevitably introduced throughout the system, including measurement noise, numerical estimation error, and coordinate transformation inaccuracies. These errors do not exist in isolation within individual modules. Instead, they propagate and accumulate along the entire localization pipeline—spanning perception, geometric reconstruction, and coordinate mapping—and are progressively amplified through multi-stage geometric estimation. As a result, such system-level error propagation can substantially degrade three-dimensional target localization accuracy and ultimately compromise the execution accuracy of the robotic manipulator.

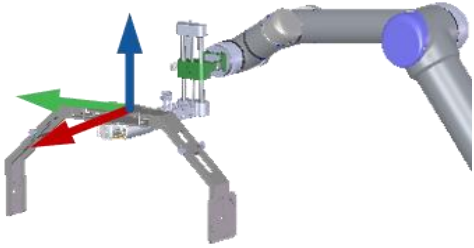

Fig.S2 Reference-based geometric perception pipeline in biplanar X-ray–guided robotic navigation. The C-arm X-ray imaging system acquires images of a localization reference structure rigidly attached to the robotic end effector. The reference structure consists of a set of fiducial points with known three-dimensional geometry. By detecting the two-dimensional pixel coordinates of these fiducials in the X-ray images and associating them with their pre-calibrated three-dimensional coordinates expressed in the connection.

As illustrated in Fig. S2, the C-arm X-ray imaging system acquires images of a localization reference structure rigidly attached to the end effector of the robotic manipulator. The reference structure contains a set of fiducial points with well-defined geometric features. By detecting the two-dimensional pixel coordinates of these fiducials in the X-ray images and associating them with their pre-calibrated three-dimensional coordinates expressed in the connector component coordinate frame, the system can estimate the transformation matrix from the connector component frame to the C-arm X-ray world coordinate frame by solving Eq. (3).

Under ideal, error-free conditions, this transformation matrix is denoted as $\mathbf{p} = A_{true} \cdot \mathbf{q}_{true}$. However, in practical systems, installation-induced structural bias between the connector component and the localization reference structure introduce systematic errors into the input three-dimensional coordinates. As a result, the estimated transformation deviates from the ideal case and is instead represented as a perturbed transformation $\mathbf{q}_{actual} = \mathbf{T}_{err1}\mathbf{q}_{true}$.

Under this condition, Eq. (1) can be rewritten as:

$$\mathbf{p} = \mathbf{A}_{actual} \cdot \mathbf{T}_{err1} \cdot \mathbf{q}_{true} \quad (8)$$

It follows directly that the estimated transformation matrix becomes $\mathbf{A}_{actual} = \mathbf{A}_{true}\mathbf{T}_{err1}^{-1}$.

Under this condition, Eq. (4) can be rewritten as $\mathbf{p}_1 = \mathbf{A}_{true}\mathbf{T}_{err1}^{-1}\mathbf{q}_{aim}$, $\mathbf{p}_2 = \mathbf{B}_{true}\mathbf{T}_{err1}^{-1}\mathbf{q}_{aim}$. Compared with $\mathbf{p}_1 = \mathbf{A} \cdot \mathbf{q}_{aim}$, $\mathbf{p}_2 = \mathbf{B} \cdot \mathbf{q}_{aim}$, we can obtain that the perturbation in $\mathbf{T}_{err1}^{-1}$ directly affects the estimation of $\mathbf{q}_{aim}$. As a consequence, the resulting three-dimensional target position is obtained as $\mathbf{q}_{L1} = \mathbf{T}_{err1}^{-1}\mathbf{q}_{aim}$.

Since the estimated position $\mathbf{q}_{L1}$ is expressed in the connector component coordinate frame, execution of the robotic motion requires transforming this position into the robot tool center point (TCP) coordinate frame.

$$\mathbf{q}_{tcp} = \mathbf{T}_{tcp}^{L2} \cdot \mathbf{T}_{L2}^{L1} \cdot \mathbf{q}_{L1} \quad (9)$$

In practice, installation-induced structural bias exist not only between the connector component and the localization reference structure, but also between the connector component and the robotic end effector. Consequently, the actual target position expressed in the TCP coordinate frame can be written as

$$\mathbf{q}_{tcp} = \mathbf{T}_{tcp}^{L2} \cdot \mathbf{T}_{L2}^{L1} \cdot \mathbf{T}_{err2} \cdot \mathbf{q}_{L1} \quad (10)$$

Under rigid-body kinematics, the installation-induced error $\mathbf{T}_{err2}$ can be decomposed into a rotational perturbation about the $x$-, $y$-, and $z$-axes and a translational offset.

$$T_{rx} = \begin{bmatrix} 1 & 0 & 0 & 0 \\ 0 & \cos\alpha & -\sin\alpha & 0 \\ 0 & \sin\alpha & \cos\alpha & 0 \\ 0 & 0 & 0 & 1 \end{bmatrix}$$

$$T_{ry} = \begin{bmatrix} \cos\beta & 0 & \sin\beta & 0 \\ 0 & 1 & 0 & 0 \\ -\sin\beta & 0 & \cos\beta & 0 \\ 0 & 0 & 0 & 1 \end{bmatrix} \quad (11)$$

$$T_{rz} = \begin{bmatrix} \cos\gamma & -\sin\gamma & 0 & 0 \\ \sin\gamma & \cos\gamma & 0 & 0 \\ 0 & 0 & 1 & 0 \\ 0 & 0 & 0 & 1 \end{bmatrix}$$

$$T_t = \begin{bmatrix} 1 & 0 & 0 & \Delta x \\ 0 & 1 & 0 & \Delta y \\ 0 & 0 & 1 & \Delta z \\ 0 & 0 & 0 & 1 \end{bmatrix} \quad (12)$$

Since the rotation angles are typically small, a first-order small-angle approximation can be applied. Specifically, the cosine terms in the transformation matrix are approximated as unity, while the sine terms are approximated by the corresponding rotation angles.

Although the small-angle approximation is adopted for analytic tractability in the first-order derivation, the Monte Carlo results demonstrate that structural amplification persists beyond the linearized regime, indicating that the observed tail-risk behavior is not an artifact of approximation.

By neglecting higher-order terms, the perturbed transformation can be expressed as:

$$\mathbf{T}_{\text{err2}} = T_t * T_{rx} * T_{ry} * T_{rz} = T = \begin{bmatrix} 1 & -\gamma & \beta & \Delta x \\ \gamma & 1 & -\alpha & \Delta y \\ -\beta & \alpha & 1 & \Delta z \\ 0 & 0 & 0 & 1 \end{bmatrix} \quad (13)$$

Under this approximation, we obtain:

$$\mathbf{q}_{\text{tcp}} = \mathbf{T}_{\text{tcp}}^{L2} \cdot \mathbf{T}_{L2}^{L1} \cdot \begin{bmatrix} 1 & -\gamma & \beta & \Delta x \\ \gamma & 1 & -\alpha & \Delta y \\ -\beta & \alpha & 1 & \Delta z \\ 0 & 0 & 0 & 1 \end{bmatrix} \cdot \mathbf{T}_{\text{err1}}^{-1} \mathbf{q}_{\text{aim}} \quad (14)$$

It follows that both $\mathbf{T}_{\text{err1}}$ and $\mathbf{T}_{\text{err2}}$ directly influence the final localization of the target point.

Importantly, these transformations modify the effective geometric configuration at which the projection and triangulation operators are evaluated. As a result, the corresponding Jacobians are functions of the biased configuration rather than the nominal geometry, thereby altering the conditioning of the estimation chain.

*C. Uncertainty and Tail-Risk Analysis*

For analytic tractability, installation-induced structural bias and pixel-level observation noise are modeled as statistically independent perturbations. This assumption isolates structural conditioning effects from statistical correlation, allowing amplification to be attributed to operator-level geometry dependence. The coupling effect analyzed in this work arises from geometry-dependent operator conditioning rather than statistical correlation.

For Eq. (3), the three-dimensional coordinates of the control points $\mathbf{q}$ and the corresponding two-dimensional image observations $\mathbf{p}$ are modeled as mutually independent random variables following zero-mean Gaussian distributions.

Let $\Sigma_{\mathbf{A}}$ denote the covariance matrix of the output $\mathbf{A}$, and let $\Sigma_{\boldsymbol{\alpha}}$ denote the covariance matrix of the input $\boldsymbol{\alpha}$. According to the principle of uncertainty propagation, the following relationship holds:

$$\mathbf{J}_{\mathbf{A}} \Sigma_{\mathbf{A}} \mathbf{J}_{\mathbf{A}}^{\top} = \mathbf{J}_{\boldsymbol{\alpha}} \Sigma_{\boldsymbol{\alpha}} \mathbf{J}_{\boldsymbol{\alpha}}^{\top} \quad (15)$$

Where $\Sigma_{\boldsymbol{\alpha}} = \text{diag}\{\Sigma_{\alpha_1},\ldots,\Sigma_{\alpha_n}\}$,

$\Sigma_{\alpha_i} = \text{diag}\{u^2(x_i), u^2(y_i), u^2(z_i), u^2(u_i), u^2(v_i)\}$,

Since the Jacobian matrices are evaluated with respect to the effective transformation matrices, installation-induced structural bias alters the Jacobian structure itself. In particular, structural bias modifies the singular value spectrum of the effective Jacobian, thereby altering error amplification characteristics across estimation stages. This change is not a uniform scaling effect, but a redistribution of sensitivity directions across the estimation chain. Consequently, structural bias modifies the mapping from measurement noise to reconstructed 3D position, rather than merely adding an independent error term.

Here, $\mathbf{J}_{\boldsymbol{\alpha}}$ denotes the Jacobian matrix of the function $G$ with respect to the input vector $\boldsymbol{\alpha}$, which consists of the three-dimensional control point coordinates x, y, z and the corresponding two-dimensional image observations u, v.

$\mathbf{J}_{\boldsymbol{\alpha}} = \text{diag}\{\mathbf{J}_{\alpha_1},\ldots,\mathbf{J}_{\alpha_n}\}$, where $\mathbf{J}_{\alpha_i}$

$$\mathbf{J}_{\alpha_i} = \begin{bmatrix} \frac{\partial G_u(\mathbf{A}, p_i, q_i)}{\partial x} & \frac{\partial G_u(\mathbf{A}, p_i, q_i)}{\partial y} & \frac{\partial G_u(\mathbf{A}, p_i, q_i)}{\partial z} & \frac{\partial G_u(\mathbf{A}, p_i, q_i)}{\partial u} & \frac{\partial G_u(\mathbf{A}, p_i, q_i)}{\partial v} \\ \frac{\partial G_v(\mathbf{A}, p_i, q_i)}{\partial x} & \frac{\partial G_v(\mathbf{A}, p_i, q_i)}{\partial y} & \frac{\partial G_v(\mathbf{A}, p_i, q_i)}{\partial z} & \frac{\partial G_v(\mathbf{A}, p_i, q_i)}{\partial u} & \frac{\partial G_v(\mathbf{A}, p_i, q_i)}{\partial v} \end{bmatrix}$$

In addition, $\mathbf{J}_{\mathbf{A}}$ denotes the Jacobian of the function $G$ with respect to the twelve parameters of the transformation matrix $\mathbf{A}$, which map the parameter vector $A_1$ to the output $A_{12}$. $\mathbf{J}_{\mathbf{A}} = [\mathbf{J}_{A_1} \cdots \mathbf{J}_{A_n}]^{\top}$,

$$\mathbf{J}_{A_i} = \begin{bmatrix} \frac{\partial G_u(\mathbf{A}, p_i, q_i)}{\partial A_1} & \cdots & \frac{\partial G_u(\mathbf{A}, p_i, q_i)}{\partial A_{12}} \\ \frac{\partial G_v(\mathbf{A}, p_i, q_i)}{\partial A_1} & \cdots & \frac{\partial G_v(\mathbf{A}, p_i, q_i)}{\partial A_{12}} \end{bmatrix}$$

We can obtain $\Sigma_{\mathbf{A}} = (\mathbf{J}_{\mathbf{A}}^{-1}) \cdot (\mathbf{J}_{\boldsymbol{\alpha}} \Sigma_{\boldsymbol{\alpha}} \mathbf{J}_{\boldsymbol{\alpha}}^{\top}) \cdot (\mathbf{J}_{\mathbf{A}}^{-1})^{\top}$.

For the computation of the three-dimensional target point coordinates, considering Eq. (5), we denote $\text{H}(\mathbf{q}_{\text{aim}}, \boldsymbol{\beta}) = 0$ as the mapping function, where $\boldsymbol{\beta} = \left[\mathbf{p}_1^{\top}, \mathbf{p}_2^{\top}, \mathbf{A}^{\top}, \mathbf{B}^{\top}\right]^{\top}$ represents the associated input variables. Let $\Sigma_{\mathbf{q}_{\text{aim}}}$ denote the covariance matrix of the output vector $\text{H}(\mathbf{q}_{\text{aim}}, \boldsymbol{\beta}) = 0$ of $\text{H}(\mathbf{q}_{\text{aim}}, \boldsymbol{\beta}) = 0$, and let $\Sigma_{\boldsymbol{\beta}}$ denote the covariance matrix of the output vector $\boldsymbol{\beta}$. According to the uncertainty propagation principle, the resulting covariance is given by $\mathbf{J}_{\mathbf{q}_{\text{aim}}} \Sigma_{\mathbf{q}_{\text{aim}}} \mathbf{J}_{\mathbf{q}_{\text{aim}}}^{\top} = \mathbf{J}_{\boldsymbol{\beta}} \Sigma_{\boldsymbol{\beta}} \mathbf{J}_{\boldsymbol{\beta}}^{\top}$.

Where

$$\mathbf{J}_{\mathbf{q}_{\text{aim}}} = \begin{bmatrix} \frac{\partial G_x(\mathbf{A}, \mathbf{p}_1, \mathbf{q}_1)}{\partial x} & \frac{\partial G_x(\mathbf{A}, \mathbf{p}_1, \mathbf{q}_1)}{\partial y} & \frac{\partial G_x(\mathbf{A}, \mathbf{p}_1, \mathbf{q}_1)}{\partial z} \\ \frac{\partial G_y(\mathbf{B}, \mathbf{p}_2, \mathbf{q}_2)}{\partial x} & \frac{\partial G_y(\mathbf{B}, \mathbf{p}_2, \mathbf{q}_2)}{\partial y} & \frac{\partial G_y(\mathbf{B}, \mathbf{p}_2, \mathbf{q}_2)}{\partial z} \end{bmatrix}$$

, $\Sigma_{\boldsymbol{\beta}} = \text{diag}\{\Sigma_{\mathbf{p}_1}, \Sigma_{\mathbf{p}_2}, \Sigma_{\mathbf{A}}, \Sigma_{\mathbf{B}}\}$,

$$\mathbf{J}_{\boldsymbol{\beta}} = \begin{bmatrix} \mathbf{J}_{\mathbf{p}_1} & 0 & \mathbf{J}_{\mathbf{A}} & 0 \\ 0 & \mathbf{J}_{\mathbf{p}_2} & 0 & \mathbf{J}_{\mathbf{B}} \end{bmatrix},$$

$$\mathbf{J}_{p_1} = \begin{bmatrix} \dfrac{\partial G_x(\mathbf{A}, p_1, \mathbf{q}_{aim})}{\partial u} & \dfrac{\partial G_x(\mathbf{A}, p_1, \mathbf{q}_{aim})}{\partial v} \\ \dfrac{\partial G_y(\mathbf{A}, p_1, \mathbf{q}_{aim})}{\partial u} & \dfrac{\partial G_y(\mathbf{A}, p_1, \mathbf{q}_{aim})}{\partial v} \end{bmatrix},$$

$$\mathbf{J}_{p_2} = \begin{bmatrix} \dfrac{\partial G_x(\mathbf{B}, p_2, \mathbf{q}_{aim})}{\partial u} & \dfrac{\partial G_x(\mathbf{B}, p_2, \mathbf{q}_{aim})}{\partial v} \\ \dfrac{\partial G_y(\mathbf{B}, p_2, \mathbf{q}_{aim})}{\partial u} & \dfrac{\partial G_y(\mathbf{B}, p_2, \mathbf{q}_{aim})}{\partial v} \end{bmatrix},$$

$$\mathbf{J}_{\mathbf{A}} = \begin{bmatrix} \dfrac{\partial G_x(\mathbf{A}, p_1, \mathbf{q}_{aim})}{\partial A_1} & \cdots & \dfrac{\partial G_x(\mathbf{A}, p_1, \mathbf{q}_{aim})}{\partial A_{12}} \\ \dfrac{\partial G_y(\mathbf{A}, p_1, \mathbf{q}_{aim})}{\partial A_1} & \cdots & \dfrac{\partial G_y(\mathbf{A}, p_1, \mathbf{q}_{aim})}{\partial A_{12}} \end{bmatrix},$$

$$\mathbf{J}_{\mathbf{B}} = \begin{bmatrix} \dfrac{\partial G_x(\mathbf{B}, p_2, \mathbf{q}_{aim})}{\partial B_1} & \cdots & \dfrac{\partial G_x(\mathbf{B}, p_2, \mathbf{q}_{aim})}{\partial B_{12}} \\ \dfrac{\partial G_y(\mathbf{B}, p_2, \mathbf{q}_{aim})}{\partial B_1} & \cdots & \dfrac{\partial G_y(\mathbf{B}, p_2, \mathbf{q}_{aim})}{\partial B_{12}} \end{bmatrix}.$$

The pseudoinverse can thus be obtained as

$$\mathbf{\Sigma}_{\mathbf{q}_{aim}} = \mathbf{J}_{\mathbf{q}_{aim}}^{-1} \mathbf{J}_{\boldsymbol{\beta}} \mathbf{\Sigma}_{\boldsymbol{\beta}} \mathbf{J}_{\boldsymbol{\beta}}^{\top} (\mathbf{J}_{\mathbf{q}_{aim}}^{\top})^{-1} \tag{16}$$

## II. EXPERIMENTS AND ANALYSIS

### A. Simulation Setup

3.2.1 Biplanar X-ray Imaging and Reconstruction

In simulation, biplanar X-ray imaging is modeled using a pinhole camera formulation. For each control point,

$$\lambda \mathbf{x}_i = \mathbf{P} \tilde{\mathbf{X}}_i, \tag{17}$$

where $\mathbf{x}_i$ denotes homogeneous pixel coordinates, $\mathbf{P}$ is the projection matrix, and $\tilde{\mathbf{X}}_i$ represents the potentially perturbed 3D control point. Projection matrices are estimated using DLT and least squares from multiple 3D–2D correspondences, and the target point is reconstructed via biplanar triangulation to obtain $\hat{\mathbf{X}}$.

3.2.2 Error Source Modeling

Two primary error sources are considered. Pixel detection noise is modeled as zero-mean Gaussian noise added to 2D observations. Installation error is treated as a structural bias, where the 3D control point input is perturbed by small rotations and translations:

$$\tilde{\mathbf{X}}_i = \mathbf{T}(\delta\theta, \delta\mathbf{t}) \mathbf{X}_i. \tag{18}$$

To reflect the execution chain in robotic navigation, the transformation from the reconstructed target to the robot TCP frame is included, with an additional installation error introduced between the mounting component and the end effector.

3.2.3 Evaluation Metrics

Performance is evaluated using the 3D localization error, TCP execution error, and average reprojection error. Tail-risk behavior is characterized using the 95th percentile (P95) and worst-case error to capture extreme failure modes arising from nonlinear geometric estimation.

### B. Simulation III: Analytic Uncertainty Propagation versus Monte Carlo

The third simulation evaluates the predictive capability of the proposed error propagation model. Under a representative operating condition, output error covariances are computed using both Monte Carlo sampling and first-order uncertainty propagation.

The comparison shows that analytic propagation accurately predicts dominant error directions and relative magnitudes, while underestimating absolute variance under strongly nonlinear conditions. This suggests that the analytic model is effective for sensitivity analysis, whereas statistical methods are required for reliable tail-risk assessment.

Under biased geometric configurations, the effective estimation operators exhibit increased curvature and conditioning shifts, which give rise to skewed and heavy-tailed execution-level error distributions. These nonlinear amplification effects cannot be fully captured by first-order linearization. The observed tail-risk dominance reflects both variance amplification and distribution skewness induced by nonlinear operator curvature under biased geometry. This confirms that amplification arises from operator-level conditioning changes rather than mere additive error accumulation. The above derivations provide the mathematical basis for the system-level structural error amplification mechanism defined in the main text.

## III. SIMULATION RESULTS

To systematically validate the proposed system-level error propagation mechanism in biplanar X-ray–guided navigation, three simulation studies are conducted to analyze:

(1) the primary effect of installation error

(2) the coupled amplification between installation error and pixel noise, and

(3) the predictive capability of first-order analytic uncertainty propagation.

All simulations are based on the biplanar pinhole imaging model described in Section III and evaluated using Monte Carlo sampling with 2000 trials per configuration. Both 3D localization error and TCP execution error are reported, together with tail-risk metrics to characterize worst-case behavior.

### A. Simulation Study I: Primary Effect of Installation Error

We first analyze the direct impact of reference structure installation error on localization and execution accuracy. Rotational and translational installation errors are independently scanned while pixel noise is fixed at a low level, so as to isolate the dominant effect of installation-induced structural bias.

*1) Rotational Installation Error*

Quantitative results are summarized in Table S1. Increasing the rotation misalignment from 0° to 2° amplifies the mean localization error from approximately 18 mm to over 75 mm. Meanwhile, the P95 tail risk of the TCP execution error nearly doubles, rising from about 45 mm to close to 90 mm.

It is important to note that, in this experiment, the TCP transformation is assumed to be an ideal rigid-body mapping without additional execution uncertainty. Consequently, the execution error magnitude matches the localization error magnitude. This observation does not imply that the execution chain is insensitive to error, but rather indicates that structural installation error introduced at the perception and reconstruction stages alone is sufficient to determine the upper bound and tail-risk behavior of navigation accuracy.

Table S1. Effect of installation rotation and translation errors on localization accuracy (Sim1)

| Installation Error | $e_{3D}$ (mm) | $e_{TCP}$ (mm) | P95 $e_{TCP}$ (mm) |
|---|---|---|---|
| Rotation 0.0° | 0.018 | 0.018 | 0.045 |
| Rotation 0.5° | 0.027 | 0.027 | 0.048 |
| Rotation 1.0° | 0.042 | 0.042 | 0.058 |
| Rotation 1.5° | 0.058 | 0.058 | 0.070 |
| Rotation 2.0° | 0.075 | 0.075 | 0.086 |
| Translation 0 mm | 0.019 | 0.019 | – |
| Translation 1 mm | 0.018 | 0.018 | – |
| Translation 2 mm | 0.019 | 0.019 | – |
| Translation 3 mm | 0.018 | 0.018 | – |
| Translation 4 mm | 0.019 | 0.019 | – |
| Translation 5 mm | 0.019 | 0.019 | – |

*2) Translational Installation Error*

As reported in Table S1, translational misalignment within the range of 0–5 mm results in only marginal variations in both localization and execution error, with error magnitudes remaining nearly constant.

*B. Simulation Study II: Coupled Amplification of Installation Error and Pixel Noise*

In practical systems, installation error and pixel-level perception noise coexist. To investigate their interaction, we perform a coupled simulation study by jointly varying installation error levels (L0–L2) and pixel noise magnitude.

This behavior is quantitatively summarized in Table S2. For example, under moderate pixel noise, the P95 TCP error increases from 90.39 mm at L0 to 108.56 mm at L2. Under high pixel noise, tail risk exceeds 200 mm regardless of whether installation error is present, indicating that the system has entered a high-risk regime.

These results demonstrate that system error is not a simple additive combination of perception noise and installation error. Instead, installation error fundamentally alters the system's sensitivity to pixel-level uncertainty, transforming otherwise tolerable noise into a navigation-level risk factor through structural coupling in the geometric estimation pipeline.

Table S2. Coupled amplification of pixel noise and installation error (P95 of execution error)

| Installation Level | $\sigma_{px} = 0\,px$ | $\sigma_{px} = 2\,px$ | $\sigma_{px} = 5\,px$ |
|---|---|---|---|
| L0 (0° + 0 mm) | 0.00 mm | 0.090 mm | 0.240 mm |
| L1 (1° + 2 mm) | 0.036 mm | 0.098 mm | 0.242 mm |
| L2 (2° + 5 mm) | 0.072 mm | 0.108 mm | 0.219 mm |

*C. Simulation Study III: Analytic Uncertainty Propagation versus Monte Carlo Statistics*

To further evaluate the predictive capability of the proposed error propagation model, we compare first-order analytic uncertainty propagation with Monte Carlo statistics under a representative operating condition.

Table S3 provides a quantitative comparison between analytic and Monte Carlo results. While the analytic model predicts the correct ordering and anisotropy of error components, the depth-direction standard deviation is underestimated by approximately a factor of two. This behavior is consistent with the linearization assumptions underlying first-order propagation and highlights its inherent limitations under strong nonlinear geometric conditions.

Rather than constituting a deficiency, this comparison clarifies the complementary roles of analytic and statistical approaches: analytic propagation provides interpretable sensitivity analysis and directional insight, while Monte Carlo sampling is necessary for reliable tail-risk estimation in nonlinear regimes.

Table S3. Comparison of analytic and Monte Carlo uncertainty propagation (Sim3)

| Axis | MC Mean (mm) | ANA Mean (mm) | MC Std (mm) | ANA Std (mm) |
|---|---|---|---|---|
| X | 0.602 | 0.602 | 0.015 | 0.010 |
| Y | 0.515 | 0.515 | 0.010 | 0.008 |
| Z | 0.221 | 0.220 | 0.043 | 0.019 |

Across the three simulation studies, the results consistently indicate that installation error acts as a primary driver of system-level error propagation in biplanar X-ray navigation. Rotational misalignment governs localization accuracy and tail-risk behavior, installation error and pixel noise exhibit coupled amplification, and first-order analytic propagation captures dominant sensitivity structure while requiring statistical evaluation for worst-case risk assessment.

## IV. REAL-SYSTEM VALIDATION

Table S4 summarizes the localization error statistics under different installation error levels. Regardless of whether 3D localization error or reprojection error is used, the error magnitude exhibits a clear monotonic increase across installation levels, i.e., L0 < L1 < L2. Under ideal installation conditions (L0), the localization error remains at a relatively low level of 0. Introducing moderate installation error (L1) increases the mean error to 0.551, while larger misalignment (L2) further degrades accuracy to 1.380. These results indicate that installation perturbations significantly deteriorate localization accuracy in a real biplanar navigation pipeline.

Beyond mean error, tail-risk metrics show an even more pronounced dependence on installation error. As the installation level increases, the growth of P95 substantially exceeds that of the mean error, indicating that installation error not only elevates overall error magnitude but also markedly amplifies worst-case behavior. This observation is consistent with the simulation results, which identified structural installation error as the dominant contributor to tail-risk amplification.

For quantitative comparison, Table S4 reports the mean error, standard deviation, and P95 values under each installation level. As shown in the table, both 3D localization error and reprojection error increase consistently with installation misalignment, with P95 being particularly sensitive to changes in installation state. This further confirms that installation error is systematically amplified through the geometric reconstruction chain in real systems.

Table S4. Real-system validation results under different installation error levels

| Level | Nominal misalignment | Mean $e_{3D}$ (mm) | Std $e_{3D}$ (mm) | P95 $e_{3D}$ (mm) | Mean $e_{reprj}$ (px) |
|---|---|---|---|---|---|
| L0 | 0.000 | 0.000 | 0.000 | 0.000 | 0.000 |
| L1 | ~1° / 2 mm | 0.551 | 0.0046 | 0.556 | 2.0 |
| L2 | ~2° / 5 mm | 1.380 | 0.0110 | 1.391 | 5.0 |